\documentclass[letterpaper]{article} 
\usepackage{aaai2027}  
\usepackage[hyphens]{url}  
\usepackage{graphicx} 
\usepackage{natbib}  
\usepackage{caption} 
\usepackage{algorithm}
\usepackage{algorithmic}
\usepackage{multirow}

\usepackage{threeparttable}
\usepackage{makecell}
\usepackage{amsmath}
\usepackage{amssymb}
\usepackage{amsfonts}
\usepackage{xcolor}
\usepackage{pifont}
\newcommand{\cmark}{\ding{51}}

\usepackage{newfloat}
\usepackage{listings}
\DeclareCaptionStyle{ruled}{labelfont=normalfont,labelsep=colon,strut=off} 
\floatstyle{ruled}
\newfloat{listing}{tb}{lst}{}
\floatname{listing}{Listing}

\usepackage{booktabs}
\title{The Next Screenshot Knows: Gated Hindsight Distillation for Mobile GUI Agents}

\author{
    Weiwei Li,
    Junzhuo Liu,
    Tong Chu, 
    Hengfu Yu,
    Wen Li\corresponding
}
\affiliations{
    University of Electronic Science and Technology of China\\
    davelee.uestc@gmail.com,
    junzhuo.cs@gmail.com,
    uestcchutong@gmail.com,
    hfyu@std.uestc.edu.cn,
    liwenbnu@gmail.com
}

\begin{document}

\maketitle

\begin{abstract}
GUI agents are commonly trained offline from successful interaction trajectories. Standard training decomposes each trajectory into prefix-action pairs: the agent predicts an action from the current screen and interaction history, while the subsequent observation is discarded. This removes the rationale of \emph{why} an action is correct: the evidence often appears only on the subsequent screen. For example, to enable \texttt{Soft Wrap}, the agent should click \texttt{Edit} or \texttt{View}, but nothing reveals this until the menu opens. Without such evidence, standard imitation gives the model little chance of ever sampling and thus learning the correct reasoning. To address this issue, we propose Gated Hindsight Distillation (GHD), which uses the next screenshot as privileged information during training. A student predicts from the observable trajectory prefix, while a parameter-sharing teacher additionally observes the next screenshot and re-scores the student’s on-policy responses. We apply distillation only when the student fails and the hindsight-conditioned teacher recovers the demonstrated action. GHD improves task success over GRPO on AndroidWorld and AndroidLab across two vision-language models. The code and checkpoints will be made available.
\end{abstract}

\section{Introduction}

\begin{figure}[!t]
    \centering
    \includegraphics[width=0.95\linewidth]{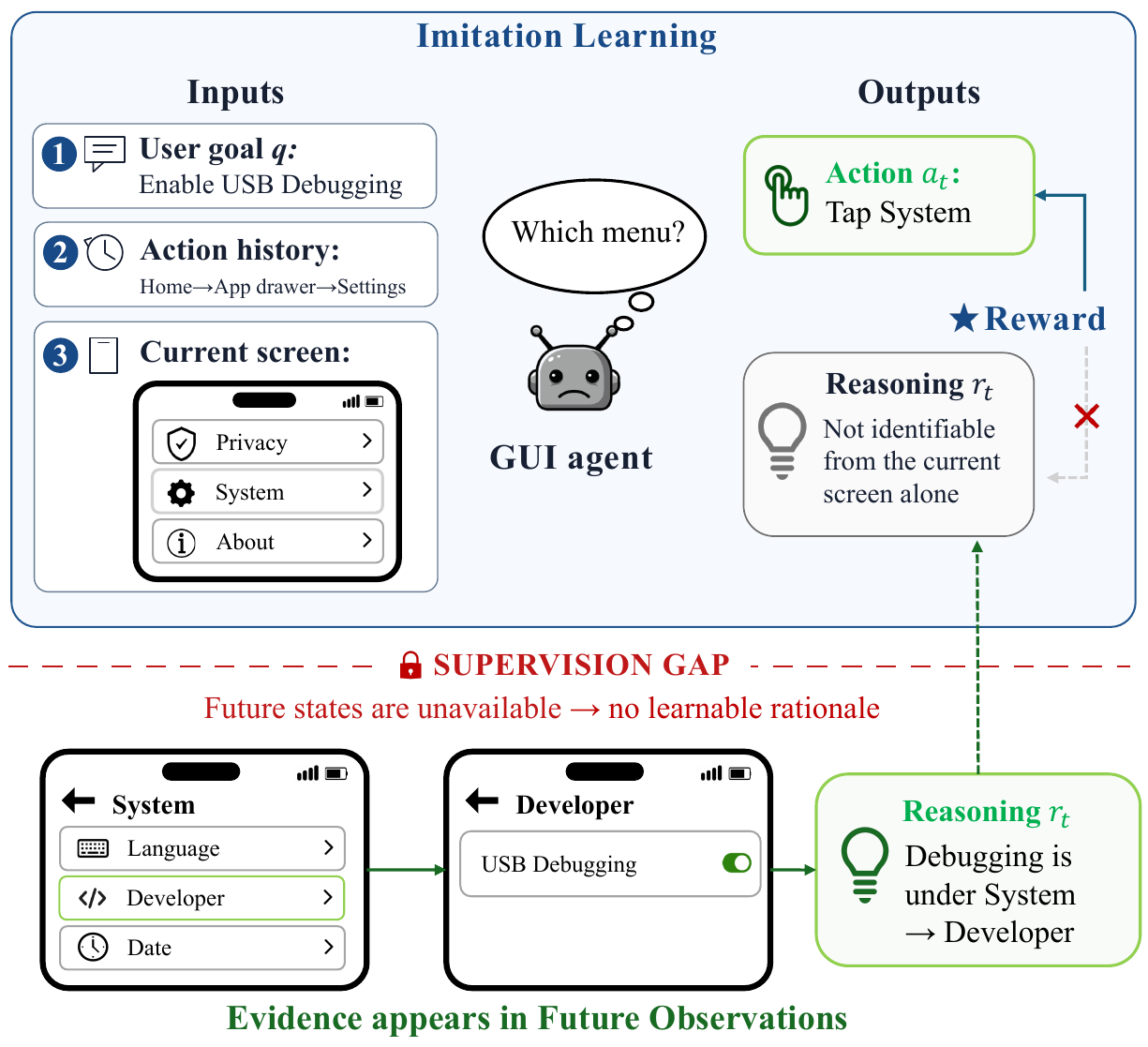}
    \caption{Supervision gap in imitation learning for GUI agents. Imitation learning directly supervises the predicted action (e.g., Tap System), while providing little explicit guidance for its reasoning. Since the evidence needed to construct an informative rationale often emerges only in future observations, the resulting rationale-learning signal is limited, making grounded action reasoning difficult to acquire through the standard imitation-learning objective alone.
    \label{fig:supervision_gap}}
    \
\end{figure}

Graphical user interface (GUI) agents operate mobile devices through the same interface as humans: they read screens and act through taps, swipes, clicks, and keystrokes. This makes them a promising solution for automating the long tail of digital tasks, including workflows that span multiple applications or expose no programmatic API \cite{wang2024guisurvey}. In practice, due to the high latency, brittle state, expensive resets, and frequently irreversible side effects of online training\cite{wu2026mobilegym,xu2026mobilerl}, most GUI agents are trained offline from previously collected successful trajectories rather than online.

To learn a generalized GUI agent that is robust to layout, task, or interaction history changes from offline demonstrations, the model must recover the app world knowledge that makes an action appropriate. In other words, it must learn grounded reasoning about the interface. Learning such reasoning requires two conditions: the training objective must \emph{supervise} the reasoning, and the model input must contain the \emph{evidence} from which that reasoning can be constructed. However, reinforcement learning based offline training methods fall short in both conditions. First, they provide no explicit supervision for reasoning. These methods typically decompose a successful trajectory into prefix--action pairs and optimize the policy to reproduce the demonstrated action~\cite{hu2026gui}. The objective teaches the policy \emph{which} action to take, but not \emph{why} it is appropriate: the application-specific rationale is never made a training target. Reasoning is therefore optimized only indirectly through action likelihood. As a result, an agent may sample a correct action while producing a plausible-sounding rationale that is not actually grounded in the interface.

Second, even if reasoning were explicitly supervised, the observable prefix often lacks the evidence needed to construct the correct rationale. Consider enabling a setting hidden under a submenu, as illustrated in Figure~\ref{fig:supervision_gap}. The current screen presents several plausible entries, and the demonstration identifies the correct click, but neither reveals the relation that the target setting lies behind that particular entry. This relation becomes observable only \emph{after} the action, when the next screenshot exposes the submenu. Future observations therefore reveal both what the action \emph{did} and why it was useful for the task. By discarding these observations, standard prefix-based training removes precisely the evidence from which a grounded rationale could be derived.

Our idea addresses both conditions at once, and it is simple: conditioning on
the next frame turns a hard prediction problem into an easy inference problem.
``What should I do?'' is answered from the prior $p(z \mid s)$, where the model
is uncertain and errs. ``What must I have done?'' is answered from the posterior
$p(z \mid s, s')$, where the answer is nearly determined: the outcome is
visible, and the rationale only needs to connect it to the goal. Reasoning drawn
from this posterior supplies the missing signal, and once written down it
becomes an explicit supervision target---the app knowledge is stated rather
than left implicit in an action label. Once stated, this knowledge can be
learned; once learned, it resolves the same ambiguity the next time it appears.

We introduce \textbf{Gated Hindsight Distillation (GHD)}, a simple framework that converts future trajectory information into supervision for a prefix-conditioned policy. During training, a teacher observes both the prefix and the future of a successful trajectory: the next observation identifies what the demonstrated action did, and the next screenshot explains how it serves the task. The resulting future-grounded rationale is distilled into a student that observes only the prefix. At inference, the future and the teacher are gone.

Future information is useful only when the teacher interprets it correctly. We therefore gate the distillation objective. A rollout is selected only when the prefix-conditioned student fails and the hindsight-conditioned teacher predicts an action matching the demonstration. This gate concentrates distillation on cases where the next screenshot provides a verifiable correction. This filters ambiguous transitions and keeps unreliable explanations out of the student's supervision.

We evaluate GHD on AndroidWorld~\cite{rawles2025androidworld} and AndroidLab~\cite{xu2025androidlab} benchmarks. Under both, GHD consistently improves task success over Supervised Fine-Tuning (SFT) and Group Relative Policy Optimization (GRPO)~\cite{guo2025deepseek}. The gains are largest where prediction is hardest: actions that require application-specific navigation knowledge or implicit prerequisites.

Our contributions are:
\begin{itemize}
    \item We identify a future-dependence problem in GUI-agent training: actions are predicted from prefixes, but the knowledge that justifies many actions is exposed only by later states.
    \item We propose Gated Hindsight Distillation, which treats the next screenshot as privileged training information and distills their knowledge into a prefix-only policy.
    \item We introduce hindsight-correction gating, which removes unreliable or ambiguous distillation targets.
    \item We demonstrate consistent improvements over baselines on AndroidWorld and AndroidLab benchmarks.
\end{itemize}

\section{Related work}

\paragraph{Offline GUI Agent Training.}
Vision-language GUI agents build on foundation models specialized for screen understanding and element grounding \citep{qin2025uitars, team2026uivenus,lin2026uivoyager}, and are typically trained offline: the policy is behavior-cloned on successful trajectories \citep{wang2024guisurvey, li2024androidcontrol, lu2025guiodyssey, chai2025amex} by supervised fine-tuning, then often refined with reinforcement learning of the GRPO family \citep{lu2025uir1, liu2025infiguir1, zhang2025agentcpmgui}. Because online interaction is costly and brittle, a productive line of work attacks data scarcity from the data side---synthesizing tasks and trajectories \citep{cheng2026openmobile}, sampling them in a hardness-aware fashion \citep{shao2026hats}, and broadening single-path logs into multi-path coverage \citep{im2025mobibench}. These methods enrich \emph{which} trajectories exist, and RL adds a sparse success signal, but each step is still supervised from its prefix alone. This leaves the future-dependence problem untouched: on the hard steps the evidence justifying an action appears only on a later screen, so the prefix is insufficient to explain---and thus to reliably learn---the correct decision.

\paragraph{Leveraging Future for GUI Agents Training.}
That the next screen carries decisive information is recognized by a growing line of GUI world models, which learn to predict the post-action UI state so an agent can plan by lookahead \citep{guan2026cuwm}, and by action-effect verification, which inspects the realized next screen to detect and recover from failed actions \citep{zhang2026verigui}. These methods consume future states at \emph{test} time---predicting or verifying them online---and pay the corresponding inference cost. GUI-Shift \cite{gao2026guishift} learns from current and future screenshots through GUI transition modeling. GHD is complementary and orthogonal: it never predicts the future. It uses the \emph{actually observed} future states of a successful demonstration purely as a training-time signal, leaving a prefix-only policy that carries no world model or verifier at deployment.

\paragraph{On-Policy Distillation for GUI Agents.}
The teacher's hindsight is transferred by distillation. Within GUI agents, distillation has been used to transfer grounding and click quality from a self- or same-context teacher \citep{huang2026trust, zhang2026learn}, to reinforce via self-distillation \citep{hubotter2026reinforcement}, and to fill off-trajectory supervision gaps with generated continuations \citep{fan2026sgcd,wu2026litegui}. The distinguishing axis of GHD is that its teacher is strictly \emph{more informed} than the student. It observes the future so distillation injects application-specific knowledge that no same-context teacher could recover from the current screen.

\begin{figure*}
    \centering
    \includegraphics[width=\linewidth]{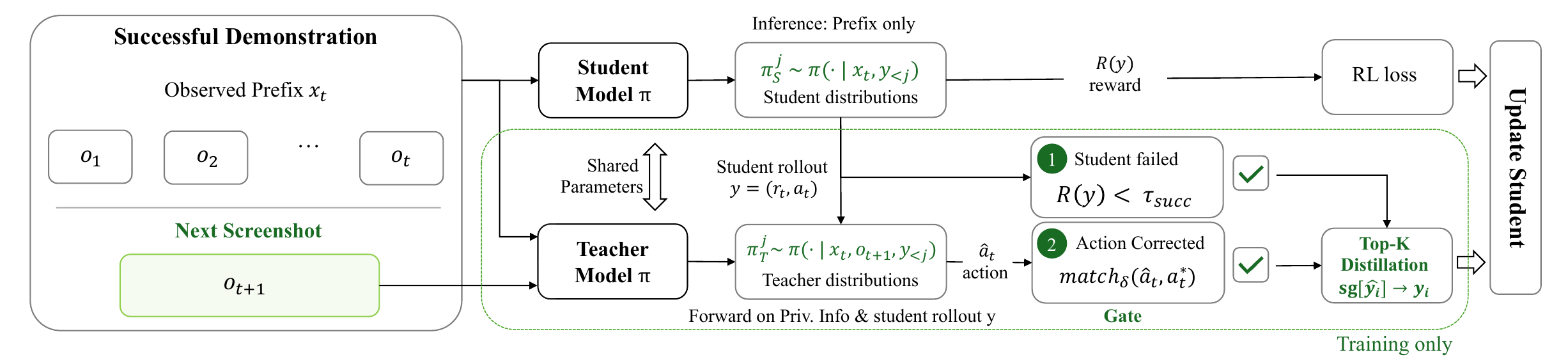}
    \caption{Overview of Gated Hindsight Distillation. Given a successful trajectory, the student observes the interaction prefix (history $o_1,\dots,o_{t-1}$ and current observation $o_t$), while a parameter-sharing teacher additionally receives the realized next screenshot $o_{t+1}$. The teacher re-scores the student's rollout under this privileged context. A gate retains distillation only when the student action fails and the teacher's position-wise correction recovers the demonstrated action. At inference, only the prefix-conditioned student is used.}
    \label{fig:overview}
\end{figure*}
\section{Method}
\label{sec:method}

We introduce \textbf{Gated Hindsight Distillation (GHD)}, which converts the
next screenshot in a successful trajectory into training-only supervision for a
prefix-conditioned GUI policy. As illustrated in Figure~\ref{fig:overview}, a
parameter-sharing teacher re-scores student rollouts with access to the next screenshot, and a gate retains the teacher signal only when it
verifiably corrects a student error. The distilled student requires neither the
future screenshot nor an additional model at inference.

\subsection{Problem Formulation}
\label{sec:problem_formulation}

A GUI task is specified by a natural-language instruction $q$. An
successful offline trajectory is
\(\tau=(q,o_1,a_1^\star,o_2,\ldots,a_{T}^\star,o_{T+1})\), where $o_t$
is the screenshot at step $t$ and $a_t^\star$ is the demonstrated action. At
decision step $t$, the causally available history is
\begin{equation}
h_t=(o_1,a_1^\star,\ldots,o_{t-1},a_{t-1}^\star,o_t),
\end{equation}
and the deployment context is $x_t=(x,h_t)$. The policy produces a response
$y_t=(r_t,a_t)$ containing a reasoning trace followed by a tool-call action:
\(\pi_\theta(y_t\mid x_t)\). During both training rollout and inference, the
student observes only $x_t$.

\subsection{Motivation}

The action label $a_t^\star$ records \emph{what} was selected but often not
\emph{why}. For example, before a menu is opened, several entries can be
visually plausible; the resulting screen reveals which entry exposes the
required control. Thus, actions must be predicted from the past, while the
evidence explaining them can appear only in the future. A scalar verifier
reward indicates whether a rollout is correct, but does not convey this
application-specific transition knowledge. In an offline successful
trajectory, however, $o_{t+1}$ is already available and directly reveals the
effect of $a_t^\star$. GHD uses $o_{t+1}$ as privileged information during
training, while preserving the causal input available at deployment.

\subsection{Reinforcement Learning}
\label{sec:rl}

At each $x_t$, the policy $\pi_{\mathrm{old}}$ samples a group of $G$ responses
\begin{equation}
y^{(n)}=(r^{(n)},a^{(n)})\sim
\pi_{\mathrm{old}}(\cdot\mid x_t),\qquad n=1,\ldots,G.
\label{eq:rollouts}
\end{equation}
We use the same step verifier for reinforcement learning and gating. For a
response $y$, $\mathrm{format}(y)=1$ iff it contains exactly one valid action, and
\(\mathrm{type}(y)=\mathbf{1}[\hat u=u^\star]\) compares the predicted and
reference action types. If the types differ, $\mathrm{value}(y)=0$.
Otherwise, the value score checks the action arguments. On the normalized
$1000\times1000$ grid, coordinate similarity is
\[
s_{\mathrm{coord}}(\hat{p},p^\star)
=
\max\left(1-\frac{\|\hat{p}-p^\star\|_2}{\sqrt{2}\cdot1000},0\right),
\]
where \texttt{click} and \texttt{long\_press} use one coordinate and
\texttt{swipe} averages its start and end coordinates. Text actions use
normalized edit similarity, while discrete arguments require an exact match.
The resulting reward is
\begin{equation}
R(y)=\tfrac12\mathrm{type}(y)+\tfrac12\mathrm{value}(y)
     +\tfrac12\mathrm{format}(y)\in[0,1.5].
\label{eq:reward}
\end{equation}
With group mean $\mu_R$ and standard deviation $\sigma_R$, GRPO assigns
\(A^{(n)}=(R(y^{(n)})-\mu_R)/(\sigma_R+\epsilon)\). Let
\(\rho_n(\theta)=\pi_\theta(y^{(n)}\mid x_t)/
\pi_{\mathrm{old}}(y^{(n)}\mid x_t)\). The clipped loss is
\begin{equation}
\begin{split}
\mathcal{L}_{\mathrm{GRPO}}=-\frac{1}{G}\sum_{n=1}^{G}\min\big(&
\rho_n(\theta)A^{(n)},\\
&\operatorname{clip}(\rho_n(\theta),1-\epsilon_c,1+\epsilon_c)A^{(n)}
\big).
\end{split}
\label{eq:grpo_objective}
\end{equation}

\subsection{Privileged Hindsight Distillation}
\label{sec:phd}

The teacher shares the student's parameters; its only advantage is the next
observation from the successful demonstration. Specifically, the student uses
$x_t$, whereas the teacher uses
\(\tilde{x}_t=(x_t,o_{t+1})\). For every sampled student response $y$, we
append its tokens to each context and run teacher forcing. At token position
$j$, the trainable student and stop-gradient teacher distributions are
\begin{equation}
\pi_S^j=\pi_\theta(\cdot\mid x_t,y_{<j}),\qquad
\pi_T^j=\operatorname{sg}\!\left[
\pi_\theta(\cdot\mid\tilde{x}_t,y_{<j})\right].
\label{eq:teacher_student}
\end{equation}
The teacher is neither separately trained nor autoregressively decoded.
Importantly, both distributions condition on the same, possibly imperfect,
student prefix $y_{<j}$. Consequently, the teacher distribution supplies dense corrections along the student's own rollout.

Following SDPO~\cite{hubotter2026reinforcement}, we minimize a generalized
Jensen--Shannon divergence
\begin{equation}
D^{(\alpha)}(\pi_T\|\pi_S)
=(1-\alpha)D_{\mathrm{KL}}(\pi_S\|m_\alpha)
+\alpha D_{\mathrm{KL}}(\pi_T\|m_\alpha),
\label{eq:jsd}
\end{equation}
where \(m_\alpha=(1-\alpha)\pi_S+\alpha\pi_T\). We use $\alpha=0.5$.
For efficiency, each divergence is evaluated on the student's top-$K$ tokens
plus one bucket containing the remaining vocabulary mass; $K=100$ in all
experiments. The gated distillation loss is
\begin{equation}
\label{eq:ghd}
\mathcal{L}_{\mathrm{GHD}}
=\mathbb{E}_{y\sim\pi_{\mathrm{old}}(\cdot\mid x_t)}
\left[M(y)\sum_{j=1}^{|y|}
D^{(\alpha)}(\pi_T^j\|\pi_S^j)\right],
\end{equation}
where $M(y)\in\{0,1\}$ is defined below. The sum covers every response token,
including both reasoning and tool-call spans, and gradients flow only through
$\pi_S$.

\subsection{Gating}
\label{sec:gating}

Privileged context is useful only when it yields a verifiable correction.
Accordingly, $M(y)$ selects responses satisfying two conditions. We define a
successful response by \(R(y)\geq\tau_{\mathrm{succ}}\). First, the
prefix-only student must fail: \(R(y)<\tau_{\mathrm{succ}}\). Second, the
hindsight-conditioned teacher must recover the demonstrated action. To test
the latter without generating a separate teacher trajectory, we take the
teacher's top-1 token at every position of the student response:
\begin{equation}
\bar y^T_j=\arg\max_v\pi_\theta(v\mid\tilde{x}_t,y_{<j}),
\qquad j=1,\ldots,|y|.
\label{eq:teacher_correction}
\end{equation}
After concatenating these position-wise predictions, we parse the resulting action as $\hat a_T$. The gate is
\begin{equation}
M(y)=\mathbf{1}\!\left[R(y)<\tau_{\mathrm{succ}}\right]\,
\mathbf{1}\!\left[\operatorname{match}_{\delta}
(\hat a_T,a_t^\star)\right].
\label{eq:gate}
\end{equation}
Both outputs must parse as JSON tool calls and have the same action
name. For \texttt{click}, \texttt{long\_press}, and \texttt{swipe}, every
coordinate must be within $\delta=20$ on the normalized grid. For
\texttt{type} and \texttt{answer}, text must match exactly or pass the
normalized edit-similarity criterion; discrete arguments must match exactly.
Missing fields, invalid values, or malformed outputs are rejected. Because
Equations~\ref{eq:teacher_student} and~\ref{eq:teacher_correction} use the same
teacher distribution conditioned on the same student prefixes, the gate
directly verifies the signal used for distillation.

\paragraph{Dynamic sampling.}
A batch may contain no student failures that the teacher corrects. For the same
prompts, we therefore draw at most three rollout-group attempts, evaluating the
gate after each attempt without updating the model. We stop at the first
attempt containing at least one accepted response; if all three fail, we retain
the last attempt. This bounds additional
generation while increasing the density of useful hindsight supervision.

\subsection{Joint Training Objective}
\label{sec:joint_objective}

GHD complements rather than replaces reinforcement learning. We optimize
\begin{equation}
\mathcal{L}=\mathcal{L}_{\mathrm{GRPO}}
+\lambda\mathcal{L}_{\mathrm{GHD}},\qquad \lambda=0.1.
\label{eq:joint_objective}
\end{equation}
The GRPO term learns from verifier reward over every rollout, whereas the GHD
term supplies token-level, future-grounded supervision only when the student
fails and the privileged teacher demonstrably corrects it. At inference, only
the prefix-conditioned policy $\pi_\theta(\cdot\mid x_t)$ is retained.

\begin{table*}[!t]
\centering
\setlength{\tabcolsep}{10pt}
\begin{threeparttable}

\begin{tabular}{l l cc cc c}
\toprule
\multirow{2}{*}{Method}
& \multirow{2}{*}{Base Model}
& \multicolumn{2}{c}{AndroidWorld}
& \multicolumn{2}{c}{AndroidLab} 
& \multicolumn{1}{c}{Average}
\\
\cmidrule(lr){3-4}\cmidrule(lr){5-6} \cmidrule(lr){7-7} 
& & Pass@1 $\uparrow$ & Pass@3 $\uparrow$
& Pass@1 $\uparrow$ & Pass@3 $\uparrow$ & Pass@1 $\uparrow$\\
\midrule

\multicolumn{6}{c}{\textit{Generic Models}} \\
\midrule
GPT-4o              & --          & 30.6 & --   &{31.2} &-- &30.8\\
Gemini-3-Pro        & --          & 60.3 & 75.0 &-- &-- &--\\

\midrule
\multicolumn{6}{c}{\textit{Open-Weight Models}} \\
\midrule
Qwen2.5-VL-7B       & --          & 25.5 & 34.9 &{10.6} &{15.2} \\
UI-Venus-7B         & Qwen2.5-VL  & 49.1 & --   &{41.3} &-- \\
Qwen3-VL-8B         & --          & 47.6 & 62.1 &{43.5} &-- \\
Step-GUI-4B         & Qwen3-VL    & 63.9 & 75.8 &{47.8} &-- \\
Step-GUI-8B         & Qwen3-VL    & 67.7 & 80.2 &-- &-- \\
MAI-UI-8B           & Qwen3-VL    & 70.7 & --   &-- &-- \\
UI-Venus-1.5-8B     & Qwen3-VL    & 73.7 & --   &{55.1} &-- \\
MobileAgent-v3.5-8B & Qwen3-VL    & 71.6 & --   &-- &-- \\

\midrule
\multicolumn{6}{c}{\textit{Open-Data Models}} \\
\midrule
UI-S1-7B            & Qwen2.5-VL  & 34.0 & --   &-- &-- & --\\
ScaleCUA-7B         & Qwen2.5-VL  & 27.2  & 36.2 &{30.0} &{37.7} & 28.6 \\
OpenMobile-7B       & Qwen2.5-VL  & 51.7  & 68.1 &{22.7} &{37.0} &37.2\\

Ours-7B                      & Qwen2.5-VL & \textbf{52.7} & 64.7 & {\textbf{43.1}} &{51.4} &\textbf{47.9}\\
\midrule
OpenMobile-8B       & Qwen3-VL    & 64.7  & 78.0 &{51.5} &{62.3} &58.1\\
Ours-8B                      & Qwen3-VL   & \textbf{66.5} & 73.3 & {\textbf{54.1}} & 65.2 &{\textbf{60.3}} \\
\bottomrule
\end{tabular}
\end{threeparttable}
\caption{
Comparison with previously reported systems on AndroidWorld and
AndroidLab. We report task-level
success as Pass@1 and Pass@3 (\%) under each system's original experimental setting. GHD achieves
the best average Pass@1 among the open-data models at both the 7B and
8B scales.
\label{tab:main_results}
}
\end{table*}
\section{Experiments}
\label{sec:experiments}

We evaluate Gated Hindsight Distillation (GHD) along five dimensions. First, we test whether training with future state as privileged information improves end-to-end task completion. Second, we conduct ablation studies to verify where the performance gain comes from. Third, how much does the next observation add? Fourth, we study the form of privileged information and the mechanism used to transfer that information to the student. Finally, how does
GHD compare with alternative uses of future GUI states?

\subsection{Experimental Setup}
\label{sec:experimental_setup}

\paragraph{Benchmarks and Metrics.}
We evaluate GHD on popular mobile GUI Agent benchmarks AndroidWorld and AndroidLab. Pass@1 denotes the task-level success rate with one rollout per task, while Pass@3 denotes success in at least one of three independent rollouts. When three evaluation rollouts are available, we report average Pass@1 over the three runs and Pass@3 as the any-success rate.

\paragraph{Models and Initialization.}
We use Qwen2.5-VL-7B~\cite{bai2025qwen25} and Qwen3-VL-8B~\cite{bai2025qwen3} as the base vision-language
models. For each backbone, we independently train an SFT checkpoint
following the OpenMobile~\cite{cheng2026openmobile} data and optimization recipe, except that
screenshots are resized to $420\times896$. This reduces visual tokens by approximately
$3\times$, improving training and
inference efficiency at the cost of a lower initial
SFT score than the published OpenMobile checkpoint.

\paragraph{GHD Implementation Details.}
All experiments use the same reproduced SFT checkpoint (denoted SFT-7B and SFT-8B), training split, rollout configuration, reward verifier, and evaluation environment for SFT, GRPO, and GHD. Unless otherwise stated, we train the models on 4 NVIDIA A100 GPUs or RTX PRO 6000 GPUs with rollout group size $G=8$, rollout temperature 1.0, maximum response length 512, and vLLM generation. The actor learning rate is $1\times10^{-6}$ with no warmup. We use GRPO without an additional KL reward penalty. For GHD, we set the distillation weight to $\lambda=0.1$, the top-$K$ logits to $K=100$, the coordinate tolerance to $\delta=20$, and train for 200 optimization steps. The experiments use the symmetric divergence $\alpha=0.5$. A sample is considered successful and excluded from distillation when its step reward exceeds $\tau_{\mathrm{succ}}=1.45$.  The teacher shares parameters with the student and differs only in receiving the privileged hint. Our final models are denoted Ours-7B and Ours-8B. The training data is a hard subset of the OpenMobile training trajectories, obtained by filtering out examples that the corresponding SFT model solves in one attempt. Starting from the full set of 27,360 training examples, this filtering yields 6,968 examples for the SFT-7B model and 5,982 examples for the SFT-8B model.

\subsection{Main Results}
\label{sec:main_results}

Table~\ref{tab:main_results} compares our models with previously
reported GUI agents\cite{gu2025ui,yan2025step,zhou2025mai,team2026uivenus,cheng2026openmobile,xu2026mobile,lu2025ui,liu2025scalecua} on AndroidWorld and AndroidLab. Since these systems differ
in model family, training data, image resolution, and algorithms,
this table is intended to establish overall competitiveness rather than
to isolate the contribution of GHD. Among open-data methods, GHD achieves the best Pass@1 at both scales, showing robustness across environment. 

\subsection{Ablation Studies}
\begin{table}[t]
\centering
\small
\setlength{\tabcolsep}{3pt}

\begin{tabular}{l c c c c c c}
\toprule
Method
& RL
& Gate
& DS
& HSD
& Pass@1 $\uparrow$
& $\Delta$ vs.\ GRPO \\
\midrule
GRPO
& \cmark & &  & &47.13 & -- \\
\quad + Gate
& \cmark  & \cmark &  && 47.84 & +0.71 \\
\quad + DS
& \cmark  & \cmark & \cmark && 49.56 & +2.43 \\
GHD
& \cmark & \cmark & \cmark& \cmark & \textbf{52.73} & \textbf{+5.60} \\
\bottomrule
\end{tabular}
\caption{
Component-wise ablation on AndroidWorld. Starting from
GRPO, \emph{+Gate} uses a teacher without privileged information to isolate
the effect of gated distillation. \emph{+DS} further enables dynamic sampling
while keeping the teacher unprivileged, controlling for the additional
rollout opportunities. Neither control receives future information. Full GHD
additionally conditions the teacher on the next screenshot and yields the
largest gain, showing that privileged future information is the primary source
of improvement. All methods start from the same SFT-7B checkpoint.
\label{tab:ablation}
}
\end{table}

\begin{table}[!t]
\centering
\small
\begin{tabular}{llcc}
\toprule
Base Model & Method
& AW Pass@1$\uparrow$ & AL Pass@1$\uparrow$ \\
\midrule
\multirow{3}{*}{Qwen2.5-VL-7B}
& SFT  & 46.55 & 29.71 \\
& GRPO & 47.13  \small $\pm$ 0.65 & 31.93 \small $\pm$ 1.12\\
& GHD  & \textbf{52.73 \small $\pm$ 1.51} & \textbf{43.10 \small $\pm$ 0.66} \\
\midrule
\multirow{3}{*}{Qwen3-VL-8B}
& SFT  & 59.05 & 39.13 \\
& GRPO & 61.35 \small $\pm$ 1.08 & 37.43 \small $\pm$ 0.42\\
& GHD  & \textbf{66.47 \small $\pm$ 0.68} & \textcolor{black}{\textbf{54.11 \small  $\pm$ 1.11}} \\
\bottomrule
\end{tabular}
\caption{Controlled comparison of SFT, GRPO, and GHD on AndroidWorld
(AW) and AndroidLab (AL). We report mean and std of Pass@1 (\%) over three independent runs. GHD shows consistent gains across
model scales and benchmarks. The best result in each comparison is bolded.
\label{tab:controlled_main_results} }

\end{table}

Table~\ref{tab:ablation} progressively controls for the non-hindsight
components of GHD. Gating contributes 0.71 points over GRPO, and dynamic
sampling increases the gain to 2.43 points. Introducing the next screenshot
then provides a further 3.17-point improvement, the largest incremental gain,
bringing full GHD to 52.73 Pass@1, showing that the majority of the improvement comes from
transferring future-grounded token-level supervision.

\subsection{Controlled Comparison}

Table~\ref{tab:controlled_main_results} compares SFT, GRPO,
and GHD under the same data, evaluation
settings, and GRPO and GHD share the same initialization. The SFT-to-GRPO difference measures the gain from
reinforcement learning, while the GRPO-to-GHD difference
isolates the contribution of hindsight distillation. For GRPO and GHD,
we report the mean and standard deviation over three independent runs. GHD shows consistent improvement over SFT and GRPO across two base models and two benchmarks.
\subsection{Privileged Information and Transfer Mechanism}
\label{sec:signal_transfer}
We next disentangle two design choices in GHD: the privileged
information available to the teacher and the mechanism used to
transfer that information to the student. All variants use the same
SFT-8B initialization, which obtains 59.05 Pass@1 on
AndroidWorld. We first hold the transfer
mechanism fixed and vary the teacher's input. We then
compare distribution-level distillation with STaR-style~\cite{2022star} self-training
under different privileged signals. 

\begin{table}[t]
\centering
\small
\setlength{\tabcolsep}{5pt}

\begin{tabular}{l c c c c c}
\toprule
Privileged Info.
& $a_t^\star$ 
& $r_t^\star$ 
& $o_{t+1}$ 
& Pass@1 $\uparrow$
& $\Delta$ vs.\ SFT \\
\midrule
SFT
&  &  &  & 59.05 & -- \\
\midrule
Distillation \\
\quad +Action
& \cmark &  &  & 58.62 & -0.43 \\
\quad +Reasoning
& \cmark & \cmark &  & 60.34 & +1.29 \\

Full & \cmark & \cmark & \cmark &64.67 & +5.62 \\
\midrule
Ours &   &   & \cmark &  \textbf{66.47} & \textbf{+7.42} \\
\bottomrule
\end{tabular}
\caption{Effect of the privileged information provided to the teacher on AndroidWorld with the 8B model. All distillation variants use the complete GHD pipeline with identical gating, dynamic sampling, rollout, and optimization settings; they differ only in the privileged information included in the teacher prompt. Providing the reference action alone does not improve over SFT,
while adding reasoning yields a modest gain. Adding next observation produces
the strongest performance gain, showing that it provides effective supervision.
\label{tab:privileged_information}}
\end{table}

Then we compare the
contribution of different privileged information. The results indicate that adding the next observation grounds predictions in the realized GUI transition and significantly boosts performance.

\paragraph{Effect of the Privileged Information.}
Table~\ref{tab:privileged_information} isolates the contribution of
the future observation while keeping the transfer mechanism fixed.
Conditioning the teacher on the demonstrated reasoning and action
improves Pass@1 from 59.05 to 60.34, a gain of 1.29 points. This
shows that access to the reference answer alone provides some useful
training signal, but the improvement is limited. Adding the next
observation $o_{t+1}$ raises performance to 64.67. The large additional improvement indicates that the next
screen contributes information not contained in the demonstrated
answer itself: it reveals the actual effect of the action and grounds
the teacher's explanation in the interface transition.

Additionally, using the screenshot alone may filter out some noisy data: offline trajectories are often synthesized using reject-sampling from exploration, which may introduce redundant steps that deviate from the user query. These steps may pass the gate by copying the reference action, while using the screenshot only  rejects them because the next screen provides no rationale for the action..

\begin{figure}
    \centering
    \includegraphics[width=0.9\linewidth]{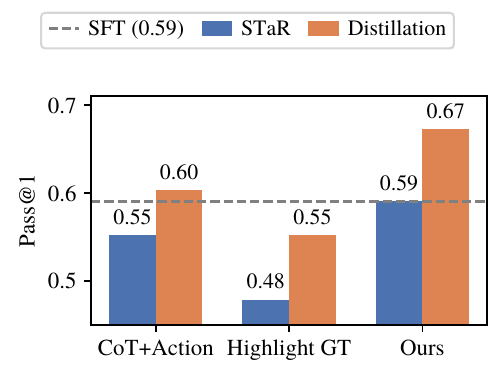}
    \caption{Effect of privileged information and transfer mechanism on
    AndroidWorld with Qwen3-VL-8B. The dashed line denotes the SFT result. We vary the teacher signal among the
    ground-truth reasoning and action (\emph{CoT+Action}), the ground-truth
    target highlighted on the current screenshot (\emph{Highlight GT}), and
    the realized next screenshot without access to the ground-truth action or
    rationale (\emph{Ours}). For each signal, STaR generates a single
    corrected reasoning-action sequence as an off-policy maximum-likelihood
    target, whereas distillation matches the teacher's token distribution
    along the student's own rollouts. Distillation consistently outperforms
    STaR across all three signals. Combining the next screenshot with
    distribution distillation performs best, showing that both the privileged signal and its
    transfer mechanism are important. }
    \label{fig:privileged_transfer}
\end{figure}

\paragraph{Different Privileged Information and Transfer Mechanisms.} We next examine whether this privileged information can be transferred
effectively using different training mechanisms. In
Figure~\ref{fig:privileged_transfer}, STaR uses the privileged teacher
to generate a single corrected reasoning--action sequence, which is
then treated as an off-policy maximum-likelihood target. Distillation,
in contrast, matches the teacher's token-level distribution along the
student's own rollouts while the student receives only
prefix information. 

Distillation outperforms STaR under every
privileged signal, suggesting that distillation provides finer-grained supervision. The form of privileged information remains important regardless of
the transfer mechanism. Highlighting the target region provides
localization information but does not explain why the target is the
appropriate prerequisite for completing the task. Revealing the
ground-truth action is more informative, but it can encourage the
teacher to copy an answer rather than infer the semantics behind it. Future observations expose the realized consequence of the
action.

\subsection{Discussion}
\label{sec:experiment_discussion}

\paragraph{Efficiency and Sampling Overhead.}
\begin{table}[t]
\centering
\small
\setlength{\tabcolsep}{4pt}
\begin{tabular}{lccc}
\toprule
Method & DS & AW Pass@1 $\uparrow$ & AL Pass@1 $\uparrow$ \\
\midrule
GRPO
&  & $61.35 \pm 1.08$ & $37.43 \pm 0.42$ \\
GHD w/o DS
&  & $63.64 \pm 2.04$ & $51.93 \pm 0.84$ \\
GHD
& \cmark & $\mathbf{66.47 \pm 0.68}$ & $\mathbf{54.11 \pm 1.11}$ \\
\bottomrule
\end{tabular}
\caption{Effect of dynamic sampling (DS) with Qwen3-VL-8B. We report
mean Pass@1 (\%) and standard deviation over three runs on AndroidWorld (AW)
and AndroidLab (AL). GRPO and GHD without DS use the same student
rollout-generation budget, with one rollout-group attempt per prompt. Full
GHD allows up to three attempts and retains the attempt containing the most
useful gated supervision. GHD without DS already outperforms GRPO, showing that its core benefit does not arise from additional
rollout generation; DS provides a complementary improvement.}
\label{tab:dynamic_sampling}
\end{table}

As shown in Table~\ref{tab:dynamic_sampling}, GHD without dynamic sampling still improves over GRPO by by 2.29 points on AndroidWorld and 14.50 points on
AndroidLab under the same generation budget. Dynamic sampling contributes a
further 2.83 and 2.18 points, respectively, by increasing the opportunity to
obtain useful gated supervision.

Dynamic sampling is bounded by three attempts and uses 2.69 attempts per batch on average over the complete training run. In our cloud training environment, enabling it did not materially increase the observed end-to-end training time relative to GHD without dynamic sampling. We treat this wall-clock observation as indicative rather than a controlled systems comparison because the experiments were conducted on non-dedicated rented instances. At deployment, GHD removes the privileged teacher, future observation, and gating procedure and therefore introduces no additional inference-time module beyond the prefix-only student.

\paragraph{Comparison with GUI-Shift-style inverse dynamics.}
Table~\ref{tab:future_information_comparison} compares GHD with a
GUI-Shift-style auxiliary task, an alternative approach for extracting
supervision from future GUI observations. GUI-Shift formulates
future-state learning as self-supervised inverse dynamics: given a
current screenshot $S_t$ and a later screenshot $S_{t+k}$, the model is
trained to predict the first action $a_t$ that initiates the observed
transition~\cite{gao2026guishift}. This objective allows the model to
learn GUI affordances and transition dynamics from trajectory data
without requiring an additional natural-language task annotation. In
our comparison, the GUI-Shift-style transition task is added as an
auxiliary objective alongside task-level GRPO starting from the same 7B SFT checkpoint.

Adding the GUI-Shift-style inverse-dynamics objective raises performance slightly to 47.41. GHD instead reaches 52.73
Pass@1. The improvement from GUI-Shift indicates
that learning to infer actions from observed state transitions provides
useful knowledge about GUI dynamics. However, its effect on the
task-conditioned policy is indirect: the inverse-dynamics task is
optimized separately and is not directly related to the main task. In contrast, GHD applies its supervision directly to the main-task prediction: the teacher's token-level distribution is distilled
on rollouts where the prefix-only student makes an error. These results
suggest that future observations are more effective when
they are converted into targeted supervision for the current decision.

\begin{table}[!t]
    \centering
    \small
    \setlength{\tabcolsep}{3pt}
    \begin{tabular}{lccc}
        \toprule
        Method &
        Transfer Mechanism &
        Pass@1 $\uparrow$ &
        $\Delta$ vs.\ GRPO \\
        \midrule
        GRPO
            & --
            & 47.13
            & -- \\
        \quad + GUI-Shift
            & Aux. Task
            & 47.41
            & +0.28 \\
        \textbf{Ours}
            & Distillation
            & \textbf{52.73}
            & \textbf{+5.60} \\
        \bottomrule
    \end{tabular}
    \caption{
        Controlled comparison of approaches for leveraging future GUI
        observations on AndroidWorld with the 7B model. All methods
        start from the same SFT checkpoint. The GUI-Shift-style
        auxiliary task presents the model with a current state and a
        future state and trains it to infer the first action that caused
        the transition. GHD instead uses the realized next observation
        as privileged teacher information and distills the resulting
        correction into a prefix-only student.
    }
    \label{tab:future_information_comparison}
\end{table}
\paragraph{Performance across AndroidLab Applications.}

Table~\ref{tab:androidlab_by_app} reports a per-application breakdown for a
representative 8B run. GHD outperforms GRPO on seven of the nine AndroidLab
applications~\cite{xu2025androidlab}, with the largest gains on Bluecoins and
Contacts; the exceptions are Calendar and Zoom, the latter containing only
five tasks.

\begin{table}[t]
\centering

\begin{tabular}{lrrrr}
\toprule
App & \# Tasks & SFT & GRPO & GHD \\
\midrule
Bluecoins & 15 & 13.33 & 26.67 & \textbf{60.00} \\
Calendar  & 14 & \textbf{50.00} & \textbf{50.00} & 42.86 \\
Cantook   & 12 & 25.00 & 25.00 & \textbf{50.00} \\
Clock     & 27 & 55.56 & 48.15 & \textbf{62.96} \\
Contacts  & 15 & 46.67 & 33.33 & \textbf{66.67} \\
Maps.me   & 15 & 6.67 & 20.00 & \textbf{33.33} \\
PiMusic   & 12 & 16.67 & 8.33 & \textbf{33.33} \\
Settings  & 23 & 65.22 & 56.52 & \textbf{73.91} \\
Zoom      & 5  & 40.00 & \textbf{60.00} & 40.00 \\
\midrule
Overall   & 138 & 39.13 & 37.68 & \textbf{55.07} \\
\bottomrule
\end{tabular}
\caption{Task success rate (\%) by AndroidLab application for a
representative 8B run. The best result in each row is bolded.}
\label{tab:androidlab_by_app}
\end{table}
\section{Conclusion}
In this work, we identified a supervision gap in the offline training of GUI agents: while models must predict actions from past and current observations, the evidence justifying these actions often appears only in future states. To address this, we introduced Gated Hindsight Distillation (GHD), a novel framework that leverages future information as a privileged training signal. By employing a future-aware teacher model, GHD extracts application-specific knowledge and distills it into a prefix-only student policy.  We also found that continuous distribution matching (distillation) is a more effective transfer mechanism than off-policy rationalization. GHD consistently improves task success on AndroidWorld and AndroidLab across different vision-language models over standard SFT and GRPO baselines. GHD provides an effective approach for offline GUI agents training, equipping them with hindsight-grounded knowledge without requiring access to future states or privileged information at deployment.  

\bibliography{aaai2027}

\end{document}